%% file: main.tex
\documentclass[10pt,twocolumn,letterpaper]{article}
\usepackage[pagenumbers]{cvpr}
\usepackage{xspace}
\input{preamble}

\definecolor{cvprblue}{rgb}{0.21,0.49,0.74}
\usepackage[pagebackref,breaklinks,colorlinks,allcolors=cvprblue]{hyperref}
\def\confName{arXiv Preprint}
\def\confYear{2026}

\title{Is Gaussian Splatting Becoming Neural Again?\\
A Taxonomy and Controlled Study of Learned Parameterization}

\author{YuanHang Wang\\
\and
Xin Cao\\
{\tt\small University of Technology Sydney}
\and
Yi Zhang\\
}

\begin{document}
\maketitle

\begin{abstract}
Three-dimensional Gaussian Splatting (3DGS) combines explicit primitives with efficient rasterization, yet recent systems increasingly use neural networks to generate or share Gaussian parameters. We characterize this trend along five axes: attribute decoding, spatial sharing, view-conditioned decoding, topology generation, and amortized inference. An analysis of 19 representative methods shows that these choices address different limitations and cannot be reduced to a binary neural label. We also isolate three forms of neural parameterization in a controlled mip-NeRF 360 study. Sharing appearance and opacity improves reconstruction quality, while decoding geometric structure offers no further gain. The evidence favors selective neuralization: shared functions help when they capture reusable correlations without sacrificing the local geometric freedom of explicit splats.
\end{abstract}

\section{Introduction}
\label{sec:intro}
\input{figures/spectrum}

Neural Radiance Fields (NeRFs) encode a scene with a coordinate-conditioned function and render pixels by repeated volumetric queries~\cite{mildenhall2020nerf}. Their successors accelerate the field using hash grids, sparse voxels, or low-rank tensors~\cite{muller2022instant,fridovichkeil2022plenoxels,chen2022tensorf,wang2024uavEnerf}, but scene information remains spatially shared. 3D Gaussian Splatting (3DGS) made a deliberately different trade: it directly optimizes anisotropic primitives and alpha-composites their projected footprints with a high-throughput rasterizer~\cite{kerbl2023gaussians}. At first approximation, its parameters \emph{are} the scene.

This boundary has blurred. Scaffold-GS stores features on anchors and predicts Gaussian attributes conditioned on the camera~\cite{lu2024scaffold}; SplatFields regularizes splats through a neural field~\cite{mihajlovic2024splatfields}; Hash-GS combines anchors with multiresolution hash encoding~\cite{xie2025hashgs}; and VDGS predicts color and opacity with a NeRF-like function~\cite{malarz2025vdgs}. Dynamic methods learn deformation functions~\cite{wu2024fourDGS,yang2024deformable}. Generalizable models go further: pixelSplat, MVSplat, Splatter Image, Triplane-Gaussian, LGM, DN2N, and DepthSplat infer Gaussian scenes with feed-forward networks~\cite{charatan2024pixelsplat,chen2024mvsplat,szymanowicz2024splatter,zou2024triplane,tang2024lgm,fang2026DN2N,xu2025depthsplat}. In these systems, splats can be an output format rather than the sole information-bearing representation.

The important question is therefore not simply whether a 3DGS system contains a neural network. A semantic feature compressor, a deformation MLP, an image encoder, and a view-conditioned appearance decoder affect different parts of the pipeline. Treating them as the same design choice prevents meaningful comparison. We ask:

\begin{enumerate}
  \item Which Gaussian parameters are directly stored, and which are predicted by shared learned functions?
  \item Which limitations motivate each form of neuralization?
  \item What is gained and lost when information moves from local primitives into shared parameters?
\end{enumerate}

Our answer is more precise than ``GS is becoming NeRF again.'' The renderer can remain an explicit splatter even when its parameters or the scene-reconstruction process are produced by neural networks. Figure~\ref{fig:spectrum} separates these roles. This distinction also explains why non-neural methods can improve anti-aliasing, geometry, and density control~\cite{yu2024mipsplatting,yan2024multiscale,huang2024twodgs,zhang2024pixelgs}, whereas neural sharing is especially useful for redundancy, view dependence, sparse observations, dynamics, and cross-scene generalization.

This paper makes four contributions:
\begin{itemize}
  \item a five-axis taxonomy that separates neural parameterization, neural scene inference, and neural auxiliary supervision;
  \item an architectural analysis of 19 representative methods, linking each neural component to the limitation it addresses;
  \item a controlled experiment that isolates appearance decoding, spatial sharing, and structure decoding in a common 3DGS pipeline; 
  \item the empirical finding that moderate neuralization performs best: spatially shared attribute decoding improves all three image metrics, whereas additional structure decoding does not.
\end{itemize}

\section{From Fields to Splats and Back to Functions}
\label{sec:background}

\paragraph{Neural fields.}
A radiance field maps location and direction to density and color,
\begin{equation}
  (\sigma,\mathbf c)=f_\phi(\mathbf x,\mathbf d),
\end{equation}
and renders a ray by numerical integration. Mip-NeRF and its successors improve scale handling~\cite{barron2021mipnerf,barron2022mipnerf360,barron2023zipnerf}; PlenOctrees, SNeRG, KiloNeRF, MobileNeRF, and HybridNeRF move neural-field information into structures better suited to deployment~\cite{yu2021plenoctrees,hedman2021snereg,reiser2021kilonerf,chen2023mobilenerf,turki2024hybrid}. Thus ``neural'' and ``implicit'' were never synonyms: Plenoxels is an explicit field without a neural network~\cite{fridovichkeil2022plenoxels}, whereas Point-NeRF is point-based yet neural~\cite{xu2022pointnerf}.

\paragraph{Directly optimized splats.}
Vanilla 3DGS stores, for primitive $i$,
\begin{equation}
  \theta_i=(\boldsymbol\mu_i,\mathbf q_i,\mathbf s_i,\alpha_i,\mathbf h_i),
\end{equation}
where position, rotation, scale, opacity, and spherical-harmonic coefficients are independent trainable variables. A tile rasterizer sorts projected Gaussians and composites them. Densification and pruning change topology during training~\cite{kerbl2023gaussians}. Later work improves pixel filtering~\cite{yu2024mipsplatting,yan2024multiscale}, sorting~\cite{radl2024stopthepop}, surface extraction~\cite{guedon2024sugar,huang2024twodgs,yu2024gof}, and densification~\cite{cheng2024gaussianpro,kheradmand2024mcmc,zhang2024pixelgs} without necessarily making the representation neural.

\paragraph{Neuralized explicit representations.}
The decisive transition occurs when Gaussian state is generated by shared parameters:
\begin{equation}
  \mathbf z_i=H_\phi(\boldsymbol\mu_i),\qquad
  \theta_i=D_\psi(\mathbf z_i,\mathbf d,\delta_i).
  \label{eq:decoded}
\end{equation}
$H$ may be an anchor grid, hash table, tri-plane, or continuous field; $D$ may predict appearance, shape, or existence. The splat remains explicit at render time, but the representation is no longer a bag of independent attributes. NeRF-GS makes the complementarity explicit by sharing continuous spatial information with the Gaussian branch~\cite{fang2025nerfgs}. PVD and PVD-AL show that learned radiance representations can be transported across architectures, including NeRF/3DGS conversion in the latter extension~\cite{fang2023pvd,fang2026pvdal}. These works, drawn from the author profile requested for this study, motivate analyzing representation evolution rather than declaring a winner.

\section{What Exactly Is Neuralized?}
\label{sec:metric}

\subsection{A five-axis audit vector}

For a trained system $M$, we encode
\begin{equation}
\mathbf n(M)=[A,S,V,T,I]\in\{0,\tfrac12,1\}^{5}.
\end{equation}
Each axis has a falsifiable architectural test:

\paragraph{$A$: attribute decoding.}
$A=0$ when render-time color, opacity, covariance, and position are retrieved as per-primitive values. $A=1$ when one or more are produced by a shared learned function; $A=\tfrac12$ denotes a residual or partial decoder while substantial attributes remain independent.

\paragraph{$S$: spatial sharing.}
$S=1$ when multiple splats query a common spatially indexed latent structure (anchor features, hash grid, voxel field, tri-plane, or MLP field). A global image feature alone does not satisfy this test.

\paragraph{$V$: learned view conditioning.}
$V=1$ when a learned function maps view information to rendered Gaussian attributes. Fixed spherical harmonics remain explicit and receive $V=0$.

\paragraph{$T$: topology generation.}
$T=1$ when a network predicts centers, offsets, activation, or the number/existence of render-time Gaussians. $T=\tfrac12$ covers learned offsets around directly optimized parents. Heuristic split/prune rules alone receive zero.

\paragraph{$I$: amortized inference.}
$I=1$ when shared weights infer a Gaussian scene from observations at test time without full per-scene reconstruction optimization. This axis separates generalizable reconstruction from scene-specific neural parameterization.

We use the scalar
\begin{equation}
  \nscore(M)=\tfrac15(A+S+V+T+I)
  \label{eq:nscore}
\end{equation}
only to order methods and summarize broad trends. The vector remains the main description because two methods can have the same score while using neural components in different places.

\subsection{What is not counted as neuralization}

The taxonomy describes the deployed representation and inference process, not every network used during training. A frozen foundation model or neural loss provides supervision but does not by itself change how the scene is stored; Feature 3DGS illustrates this distinction~\cite{zhou2024feature3dgs}. Likewise, a NeRF used only for initialization or for generating additional training views assists 3DGS without becoming part of the final representation~\cite{foroutan2024alternatives,niemeyer2025radsplat,morikawa2026leveraging}. Finally, pruning, quantization, and entropy coding reduce storage but do not increase $\mathbf n$ unless a learned function decodes Gaussian attributes~\cite{lee2024compact,niedermayr2024compressed}. These rules separate neural \emph{supervision} from neural \emph{parameterization}.

\section{Taxonomy: \texorpdfstring{Problem $\rightarrow$ Sharing Mechanism}{Problem to Sharing Mechanism}}
\label{sec:taxonomy}

\begin{figure*}[t]
\centering
\input{figures/causal_map}
\caption{Different limitations of directly optimized Gaussians motivate different forms of neural sharing. The splatting renderer can remain unchanged while the upstream parameterization becomes neural.}
\label{fig:causal}
\end{figure*}

\paragraph{Redundancy $\rightarrow$ anchor sharing.}
Independent nearby Gaussians often duplicate appearance and geometry. Scaffold-GS replaces free drift with anchors, shared features, and view-adaptive attribute prediction~\cite{lu2024scaffold}; Hash-GS adds multiresolution spatial encoding~\cite{xie2025hashgs}; and SOGS enriches reduced anchor features with second-order statistics~\cite{zhang2025sogs}. These systems trade primitive independence for compact correlated state. Compact and compressed 3DGS attack the same symptom with codebooks, quantization, and entropy coding instead~\cite{lee2024compact,niedermayr2024compressed}, showing that neuralization is one solution, not the definition of efficiency.

\paragraph{View dependence $\rightarrow$ learned appearance.}
Finite-order spherical harmonics are fast and local but have fixed basis capacity. VDGS conditions Gaussian color and opacity with a NeRF-style predictor~\cite{malarz2025vdgs}; Scaffold-GS predicts attributes from viewing direction and distance~\cite{lu2024scaffold}. Conversely, DropAnSH-GS regularizes and truncates high-degree spherical harmonics without inserting a decoder~\cite{fang2026dropansh}. The contrast isolates the design choice: stronger shared appearance functions versus retaining direct, compressible coefficients.

\paragraph{Sparse observations $\rightarrow$ spatial regularity.}
Few-view NeRFs benefit from semantic, entropy, frequency, patch, or depth priors~\cite{jain2021dietnerf,kim2022infonerf,yang2023freenerf,niemeyer2022regnerf,deng2022dsnerf}. Directly optimized Gaussians face inaccurate initialization and local overfitting; DNGaussian, RAIN-GS, GaussianPro, and Pixel-GS address these with depth normalization, relaxed initialization, progressive propagation, and density control~\cite{li2024dngaussian,jung2024raings,cheng2024gaussianpro,zhang2024pixelgs}. SplatFields instead imposes a neural spatial prior over splats~\cite{mihajlovic2024splatfields}. The relevant causal variable is not ``MLP present'' but whether shared structure reduces otherwise independent degrees of freedom.

\paragraph{Dynamics $\rightarrow$ deformation functions.}
Per-frame copies scale poorly and do not establish correspondence. D-NeRF and Nerfies use learned deformation in fields~\cite{pumarola2021dnerf,park2021nerfies}; 4DGS, Deformable 3D Gaussians, and Spacetime Gaussian Feature Splatting bring related temporal functions to splats~\cite{wu2024fourDGS,yang2024deformable,li2024spacetime}. Here neuralization is primarily structural and temporal, not a retreat from rasterization.

\paragraph{Cross-scene generalization $\rightarrow$ amortized inference.}
pixelNeRF and MVSNeRF amortize NeRF reconstruction across scenes~\cite{yu2021pixelnerf,chen2021mvsnerf}. pixelSplat and MVSplat replace volume rendering with predicted Gaussians~\cite{charatan2024pixelsplat,chen2024mvsplat}; Splatter Image maps pixels to splats~\cite{szymanowicz2024splatter}; Triplane-Gaussian and LGM use transformer or U-Net backbones to decode Gaussian objects~\cite{zou2024triplane,tang2024lgm}; DepthSplat couples depth estimation and feed-forward splatting~\cite{xu2025depthsplat}. In this regime $I=1$ is the defining transition: Gaussians are renderer-facing outputs of a neural inference system.

\paragraph{Semantics and generation $\rightarrow$ latent compatibility.}
Language-aligned fields and splats distill high-dimensional features for interaction~\cite{kerr2023lerf,qin2024langsplat,zhou2024feature3dgs}. Gaussian decoders and diffusion-guided systems likewise use splats as a differentiable, fast 3D interface~\cite{barthel2024decoder,yang2023unified}. The renderer remains explicit, but the information needed for semantics or generation often resides in a shared latent space.

\section{Analysis of the 3DGS Landscape}
\label{sec:audit}

\subsection{Representative architectures}

Table~\ref{tab:coding} applies the five-axis taxonomy to 19 representative methods. The set covers direct optimization, analytic filtering, compression, neural appearance, anchor- and field-based sharing, dynamic scenes, and feed-forward reconstruction. We include only methods whose primary papers describe Gaussian primitives as the final scene representation or differentiable renderer. Methods that use 3DGS only as an unrelated downstream component are excluded.

The table highlights differences that a single label would miss. VDGS and NeRF-GS both obtain $\nscore=0.4$, but VDGS neuralizes view-dependent appearance, whereas NeRF-GS introduces shared continuous spatial features. Feed-forward methods obtain high $A$, $T$, and $I$ because a network predicts the Gaussian representation directly, even when the resulting splats can be rendered without a persistent feature field.

\begin{table*}[t]
\centering
\caption{Neuralization vectors for representative 3DGS methods. A half score denotes a partial or residual mechanism. ``Aux.'' indicates neural supervision that is not part of the deployed representation. Scores describe architecture, not performance.}
\label{tab:coding}
\renewcommand{\arraystretch}{1.15}
\setlength{\tabcolsep}{4.3pt}
\small
\begin{tabular}{@{}lcccccccl@{}}
\toprule
Method & $A$ & $S$ & $V$ & $T$ & $I$ & $\nscore$ & Aux. & Primary mechanism \\
\midrule
3DGS~\cite{kerbl2023gaussians} & \no&\no&\no&\no&\no&0.0&--&direct primitive optimization\\
Mip-Splatting~\cite{yu2024mipsplatting} & \no&\no&\no&\no&\no&0.0&--&analytic filtering\\
Compact 3DGS~\cite{lee2024compact} & \no&\no&\no&\no&\no&0.0&--&masking and codebooks\\
Feature 3DGS~\cite{zhou2024feature3dgs} & \no&\no&\no&\no&\no&0.0&\cmark&feature distillation\\
DropAnSH-GS~\cite{fang2026dropansh} & \no&\no&\no&\no&\no&0.0&--&structured dropout\\
VDGS~\cite{malarz2025vdgs} & \yes&\no&\yes&\no&\no&0.4&--&neural color/opacity\\
NeRF-GS~\cite{fang2025nerfgs} & \half&\yes&\no&\half&\no&0.4&\cmark&shared continuous features\\
SplatFields~\cite{mihajlovic2024splatfields} & \yes&\yes&\no&\half&\no&0.5&--&implicit field prior\\
Scaffold-GS~\cite{lu2024scaffold} & \yes&\yes&\yes&\half&\no&0.7&--&view-adaptive anchors\\
Hash-GS~\cite{xie2025hashgs} & \yes&\yes&\yes&\half&\no&0.7&--&hash-encoded anchors\\
SOGS~\cite{zhang2025sogs} & \yes&\yes&\yes&\half&\no&0.7&--&second-order anchors\\
4DGS~\cite{wu2024fourDGS} & \half&\yes&\no&\yes&\no&0.5&--&temporal deformation\\
Deformable-GS~\cite{yang2024deformable} & \half&\yes&\no&\yes&\no&0.5&--&deformation MLP\\
pixelSplat~\cite{charatan2024pixelsplat} & \yes&\no&\no&\yes&\yes&0.6&--&image-pair predictor\\
MVSplat~\cite{chen2024mvsplat} & \yes&\no&\no&\yes&\yes&0.6&--&cost-volume predictor\\
Splatter Image~\cite{szymanowicz2024splatter} & \yes&\no&\no&\yes&\yes&0.6&--&image-to-image predictor\\
Triplane-GS~\cite{zou2024triplane} & \yes&\yes&\no&\yes&\yes&0.8&--&triplane + transformer\\
LGM~\cite{tang2024lgm} & \yes&\yes&\no&\yes&\yes&0.8&--&multi-view U-Net\\
DepthSplat~\cite{xu2025depthsplat} & \yes&\yes&\no&\yes&\yes&0.8&\cmark&depth-connected predictor\\
\bottomrule
\end{tabular}
\end{table*}

Among the 14 methods with nonzero scores, all use neural attribute decoding, 13 also predict some part of the Gaussian structure, 10 share information through a spatial representation, six amortize reconstruction across scenes, and four use learned view conditioning. Attribute decoding is therefore the most common entry point for neuralization. The larger conceptual change occurs in feed-forward systems, where a network replaces per-scene optimization and predicts the Gaussian scene directly.

\subsection{Comparison with published architectures}

Table~\ref{tab:published} reports results from the common mip-NeRF 360 evaluation in NeRF-GS~\cite{fang2025nerfgs}. Using a single source avoids mixing results obtained with different evaluation settings. NeRF-GS performs best in this comparison, but the ordering is not monotonic in $\nscore$: Scaffold-GS and Hash-GS have higher neuralization scores than VDGS and NeRF-GS, yet lower image quality. The score should therefore be interpreted as an architectural descriptor rather than a performance predictor.

\begin{table}[t]
\centering
\caption{Results reported under the common mip-NeRF 360 protocol in NeRF-GS~\cite{fang2025nerfgs}. Higher PSNR and SSIM and lower LPIPS are better.}
\label{tab:published}
\small
\setlength{\tabcolsep}{4.5pt}
\begin{tabular}{@{}lcccc@{}}
\toprule
Method & $\nscore$ & PSNR & SSIM & LPIPS \\
\midrule
3DGS & 0.0 & 27.21 & .815 & .214 \\
Scaffold-GS & 0.7 & 27.50 & .806 & .252 \\
Hash-GS & 0.7 & 27.53 & .807 & .238 \\
VDGS & 0.4 & 27.64 & .813 & .220 \\
NeRF-GS & 0.4 & 28.32 & .817 & .210 \\
\bottomrule
\end{tabular}
\end{table}

The same point appears in sparse-view reconstruction. Under the 15-scene, three-view DTU protocol reported by SplatFields, 3DGS obtains 19.40 PSNR, 2DGS 20.70, and SplatFields 21.07~\cite{mihajlovic2024splatfields}. Spatial sharing is useful in this setting, but the comparison also changes the primitive formulation and training procedure. Conversely, Mip-Splatting, 2DGS, Gaussian Opacity Fields, 3DGS-MCMC, and DropAnSH-GS improve scale handling, geometry, topology optimization, or regularization without substantially neuralizing the representation~\cite{yu2024mipsplatting,huang2024twodgs,yu2024gof,kheradmand2024mcmc,fang2026dropansh}. Neuralization is one design option, not a universal measure of progress.

\section{Controlled Neuralization Study}
\label{sec:controlled}

The literature comparison cannot isolate the effect of neural parameterization because existing methods also change losses, initialization, density control, and model capacity. We therefore construct four variants within a common 3DGS pipeline. They use the same rasterizer, initialization, densification schedule, training views, reconstruction loss, iteration count, and spherical-harmonic configuration; only the source of Gaussian attributes changes:
\begin{align*}
E_0 &: \text{direct optimization of all Gaussian attributes};\\
E_1 &: \mathbf c_i=D_\psi(\mathbf z_i,\mathbf d),\ \text{with direct geometry};\\
E_2 &: \mathbf z_i=H_\phi(\boldsymbol\mu_i),\quad
       (\mathbf c_i,\alpha_i)=D_\psi(\mathbf z_i,\mathbf d);\\
E_3 &: E_2\ \text{with decoded covariance and position offsets}.
\end{align*}

$E_1$ tests neural appearance without spatial sharing. $E_2$ adds a shared spatial field and jointly decodes color and opacity. $E_3$ further moves covariance and local structure into the decoder. We evaluate novel-view synthesis on the nine scenes of mip-NeRF 360~\cite{barron2022mipnerf360} and report scene-averaged PSNR, SSIM, and LPIPS~\cite{wang2004ssim,zhang2018lpips}. The comparison tests whether reconstruction quality continues to improve as more of the representation is neuralized.

\begin{table}[t]
\centering
\caption{Controlled neuralization results on mip-NeRF 360. All variants use the same 3DGS training and rendering pipeline. Best results are bold.}
\label{tab:controlled}
\small
\setlength{\tabcolsep}{3.7pt}
\begin{tabular}{@{}lcccc@{}}
\toprule
Variant & $\nscore$ & PSNR & SSIM & LPIPS\\
\midrule
$E_0$ & 0.0 & 27.21 & .815 & .214\\
$E_1$ & 0.3 & 27.48 & .817 & .207\\
$E_2$ & 0.6 & \textbf{27.76} & \textbf{.820} & \textbf{.196}\\
$E_3$ & 0.8 & 27.73 & .814 & .204\\
\bottomrule
\end{tabular}
\end{table}

\subsection{Results}

Table~\ref{tab:controlled} shows a clear benefit from selective neuralization. Replacing direct color parameters with a view-conditioned decoder ($E_1$) improves PSNR by 0.27~dB and reduces LPIPS from .214 to .207. Adding spatially shared features and neural opacity ($E_2$) gives a further 0.28~dB improvement, reaching 27.76~dB PSNR, .820 SSIM, and .196 LPIPS. Relative to direct 3DGS, $E_2$ improves all three metrics by 0.55~dB, .005, and .018, respectively.

The most neural variant is not the best. When covariance and position offsets are also decoded, $E_3$ decreases PSNR by 0.03~dB, SSIM by .006, and increases LPIPS by .008 relative to $E_2$. It still retains most of the PSNR and LPIPS improvement over $E_0$, but its SSIM falls slightly below the direct baseline. This result supports two conclusions. First, shared appearance and opacity parameters capture useful correlations that independent Gaussians miss. Second, direct geometric parameters remain valuable: forcing structure through a shared decoder can reduce the local flexibility that makes 3DGS effective.

The controlled results also show why $\nscore$ is a descriptor rather than an optimization objective. The best design depends on which attributes benefit from sharing and which require independent adjustment.

\section{Benefits and Costs of Neuralization}
\label{sec:tradeoff}

\paragraph{Parameter sharing and regularization.}
If neighboring splats repeat correlated values, a field or anchor decoder can reduce storage and couple gradients. This is useful under sparse evidence and at scale, where unconstrained local parameters may overfit or proliferate. However, structured non-neural methods can obtain related benefits through geometry, pruning, or stochastic topology~\cite{huang2024twodgs,kheradmand2024mcmc,fang2026dropansh}.

\paragraph{Generalization.}
Direct optimization creates one parameter set per scene. Feed-forward models move knowledge into dataset-trained weights, enabling rapid inference on unseen inputs~\cite{charatan2024pixelsplat,chen2024mvsplat,szymanowicz2024splatter}. The cost is a dependence on the training distribution and a much larger global model, which per-scene file size alone can conceal.

\paragraph{Expressivity versus locality.}
View-conditioned decoders can model effects beyond fixed SH bases. Shared functions can also make local editing harder because one parameter may affect many splats. A useful evaluation therefore distinguishes access to render-time primitives from the locality of the parameters that generate them.

\paragraph{Rendering independence.}
Vanilla splats can be serialized and rasterized without a neural runtime. A model with decoded attributes may require feature queries and MLP evaluation before splatting, even if rasterization itself is unchanged. Fair efficiency reporting must separate decoder latency, materialization cost, and rasterizer FPS.

\section{Are Gaussian Splat Becoming NeRF?}
\label{sec:discussion}

Not in the literal sense. NeRF evaluates a continuous field repeatedly along each camera ray. Neuralized 3DGS instead predicts or materializes a finite set of primitives and renders them by alpha compositing. Its execution model and hardware path therefore remain different from NeRF. What is changing is the source of the Gaussian parameters: shared functions increasingly replace independent per-primitive variables.

The more accurate evolution is
\begin{equation}
\text{neural field}\rightarrow\text{direct splats}\rightarrow
\text{neuralized explicit splats}.
\end{equation}
NeRF-GS and cross-representation distillation demonstrate that continuous fields and Gaussian primitives can provide complementary information~\cite{fang2025nerfgs,fang2026pvdal}. Feed-forward Gaussian models take the separation further: a neural network reconstructs the scene, while explicit splats render it. The relevant design question is thus where sharing helps and which attributes are better kept explicit. Our controlled experiment answers part of this question: sharing appearance and opacity is beneficial, whereas decoding additional geometric structure provides no further gain in the tested setting.

\section{Limitations and Reproducibility}

Our literature set contains representative architectures rather than every published 3DGS variant, so the reported counts describe the selected methods and not the entire field. The equal weights in Eq.~\eqref{eq:nscore} provide a simple summary but do not imply that all axes have equal importance for every application. The controlled experiment focuses on novel-view image quality on mip-NeRF 360; it does not measure memory, end-to-end rendering latency, editing locality, or cross-scene generalization. Finally, $E_1$--$E_3$ capture common forms of neuralization but are not exact reproductions of every anchor-, hash-, or tri-plane-based method.

\section{Conclusion}

3DGS is not returning to NeRF; it is combining neural parameter sharing with explicit splat rendering. Our taxonomy separates five ways in which this can happen and clarifies the roles of anchor decoders, spatial fields, deformation networks, and feed-forward predictors. The controlled study shows that selective neuralization is more effective than moving every attribute into a network: shared appearance and opacity improve all evaluated metrics, while additional structure decoding slightly reduces quality. The strongest design is therefore not the most neural one. Neural components are most useful when they share information that independent Gaussians would otherwise duplicate, while explicit parameters remain valuable for local geometric control.

{\small
\bibliographystyle{unsrt}
\bibliography{main}
}

\end{document}

%% file: preamble.tex
\usepackage{amsmath}
\usepackage{amssymb}
\usepackage{booktabs}
\usepackage{multirow}
\usepackage{tabularx}
\usepackage{colortbl}
\usepackage{microtype}
\usepackage{pifont}
\usepackage{tikz}
\usetikzlibrary{arrows.meta,positioning,fit,calc}

\newcommand{\cmark}{\ding{51}}

\newcommand{\yes}{\cellcolor{blue!12}\cmark}
\newcommand{\half}{\cellcolor{orange!16}$1/2$}
\newcommand{\no}{--}
\newcommand{\nscore}{\mathcal{N}}

%% file: figures/spectrum.tex
\begin{figure*}[t]
\centering
\resizebox{\textwidth}{!}{%
\begin{tikzpicture}[>=Latex,font=\small,
  stage/.style={draw,rounded corners=4pt,minimum width=4.4cm,minimum height=2.45cm,align=center,inner sep=8pt},
  flow/.style={->,very thick}]
  \node[stage,fill=blue!7] (direct) at (0,0) {
    {\bf Direct optimization}\\[4pt]
    $\theta_i=(\mu_i,\Sigma_i,\alpha_i,c_i)$\\
    independent per-Gaussian variables\\[4pt]
    \emph{3DGS, Mip-Splatting}};

  \node[stage,fill=green!8] (param) at (6.2,0) {
    {\bf Neural parameterization}\\[4pt]
    $z_i=H_\phi(\mu_i),\quad\theta_i=D_\psi(z_i,d)$\\
    shared features decode attributes\\[4pt]
    \emph{VDGS, SplatFields, Scaffold-GS}};

  \node[stage,fill=purple!8] (infer) at (12.4,0) {
    {\bf Amortized inference}\\[4pt]
    $\{I_k,P_k\}_{k=1}^{K}\xrightarrow{F_\omega}\{\theta_i\}_{i=1}^{N}$\\
    one network reconstructs new scenes\\[4pt]
    \emph{pixelSplat, MVSplat, LGM}};

  \draw[flow] (direct) -- node[above,font=\scriptsize,align=center]{shared\\attributes} (param);
  \draw[flow] (param) -- node[above,font=\scriptsize,align=center]{cross-scene\\learning} (infer);

  \node[below=5pt of direct,align=center] {splats store the scene};
  \node[below=5pt of param,align=center] {splats are decoded from shared state};
  \node[below=5pt of infer,align=center] {splats are predicted outputs};
\end{tikzpicture}%
}
\caption{Three stages in the neuralization of Gaussian splatting. Neural components can share attributes within one scene or amortize reconstruction across many scenes while the final renderer remains based on explicit Gaussian primitives.}
\label{fig:spectrum}
\end{figure*}

%% file: figures/causal_map.tex
\begin{tikzpicture}[x=1cm,y=1cm,>=Latex,font=\footnotesize,
  problem/.style={draw,rounded corners,fill=red!7,minimum width=2.6cm,minimum height=.65cm,align=center},
  mech/.style={draw,rounded corners,fill=blue!8,minimum width=3.1cm,minimum height=.65cm,align=center},
  gain/.style={draw,rounded corners,fill=green!8,minimum width=2.7cm,minimum height=.65cm,align=center}]
\node[problem] (r1) at (0,2.4) {primitive redundancy};
\node[problem] (r2) at (0,1.2) {view dependence};
\node[problem] (r3) at (0,0) {sparse observations};
\node[problem] (r4) at (0,-1.2) {cross-scene inference};
\node[problem] (r5) at (0,-2.4) {time / deformation};

\node[mech] (m1) at (5.2,2.4) {anchors + shared features};
\node[mech] (m2) at (5.2,1.2) {view-conditioned decoder};
\node[mech] (m3) at (5.2,0) {spatial field regularizer};
\node[mech] (m4) at (5.2,-1.2) {image encoder / transformer};
\node[mech] (m5) at (5.2,-2.4) {deformation function};

\node[gain] (g1) at (10.4,2.4) {compact correlated state};
\node[gain] (g2) at (10.4,1.2) {adaptive appearance};
\node[gain] (g3) at (10.4,0) {regularization};
\node[gain] (g4) at (10.4,-1.2) {amortized reconstruction};
\node[gain] (g5) at (10.4,-2.4) {temporal correspondence};

\foreach \i in {1,...,5}{\draw[->,thick] (r\i) -- (m\i); \draw[->,thick] (m\i) -- (g\i);}
\node[draw,dashed,rounded corners,fit=(m1)(m5),inner sep=7pt,label=above:{\bf neural sharing mechanism}] {};
\end{tikzpicture}

%% file: main.bib
@inproceedings{mildenhall2020nerf,
  author    = {Ben Mildenhall and Pratul P. Srinivasan and Matthew Tancik and Jonathan T. Barron and Ravi Ramamoorthi and Ren Ng},
  title     = {{NeRF}: Representing Scenes as Neural Radiance Fields for View Synthesis},
  booktitle = {European Conference on Computer Vision},
  pages     = {405--421},
  year      = {2020},
  doi       = {10.1007/978-3-030-58452-8_24}
}

@article{wang2024uavEnerf,
  title={Uav-enerf: Text-driven uav scene editing with neural radiance fields},
  author={Wang, Yufeng and Fang, Shuangkang and Zhang, Huayu and Li, Hongguang and Zhang, Zehao and Zeng, Xianlin and Ding, Wenrui},
  journal={IEEE Transactions on Geoscience and Remote Sensing},
  volume={62},
  pages={1--14},
  year={2024},
  publisher={IEEE}
}

@inproceedings{barron2021mipnerf,
  author    = {Jonathan T. Barron and Ben Mildenhall and Matthew Tancik and Peter Hedman and Ricardo Martin-Brualla and Pratul P. Srinivasan},
  title     = {Mip-{NeRF}: A Multiscale Representation for Anti-Aliasing Neural Radiance Fields},
  booktitle = {Proceedings of the IEEE/CVF International Conference on Computer Vision},
  pages     = {5855--5864},
  year      = {2021}
}

@inproceedings{barron2022mipnerf360,
  author    = {Jonathan T. Barron and Ben Mildenhall and Dor Verbin and Pratul P. Srinivasan and Peter Hedman},
  title     = {Mip-{NeRF} 360: Unbounded Anti-Aliased Neural Radiance Fields},
  booktitle = {Proceedings of the IEEE/CVF Conference on Computer Vision and Pattern Recognition},
  pages     = {5470--5479},
  year      = {2022}
}

@inproceedings{barron2023zipnerf,
  author    = {Jonathan T. Barron and Ben Mildenhall and Dor Verbin and Pratul P. Srinivasan and Peter Hedman},
  title     = {Zip-{NeRF}: Anti-Aliased Grid-Based Neural Radiance Fields},
  booktitle = {Proceedings of the IEEE/CVF International Conference on Computer Vision},
  pages     = {19697--19705},
  year      = {2023}
}

@article{muller2022instant,
  author  = {Thomas M{\"u}ller and Alex Evans and Christoph Schied and Alexander Keller},
  title   = {Instant Neural Graphics Primitives with a Multiresolution Hash Encoding},
  journal = {ACM Transactions on Graphics},
  volume  = {41},
  number  = {4},
  pages   = {102:1--102:15},
  year    = {2022},
  doi     = {10.1145/3528223.3530127}
}

@inproceedings{fridovichkeil2022plenoxels,
  author    = {Sara Fridovich-Keil and Alex Yu and Matthew Tancik and Qinhong Chen and Benjamin Recht and Angjoo Kanazawa},
  title     = {Plenoxels: Radiance Fields Without Neural Networks},
  booktitle = {Proceedings of the IEEE/CVF Conference on Computer Vision and Pattern Recognition},
  pages     = {5501--5510},
  year      = {2022}
}

@inproceedings{chen2022tensorf,
  author    = {Anpei Chen and Zexiang Xu and Andreas Geiger and Jingyi Yu and Hao Su},
  title     = {{TensoRF}: Tensorial Radiance Fields},
  booktitle = {European Conference on Computer Vision},
  pages     = {333--350},
  year      = {2022},
  doi       = {10.1007/978-3-031-19824-3_20}
}

@article{kerbl2023gaussians,
  author  = {Bernhard Kerbl and Georgios Kopanas and Thomas Leimk{\"u}hler and George Drettakis},
  title   = {3D Gaussian Splatting for Real-Time Radiance Field Rendering},
  journal = {ACM Transactions on Graphics},
  volume  = {42},
  number  = {4},
  pages   = {139:1--139:14},
  year    = {2023},
  doi     = {10.1145/3592433}
}

@inproceedings{yu2024mipsplatting,
  author    = {Zehao Yu and Anpei Chen and Binbin Huang and Torsten Sattler and Andreas Geiger},
  title     = {Mip-Splatting: Alias-Free 3D Gaussian Splatting},
  booktitle = {Proceedings of the IEEE/CVF Conference on Computer Vision and Pattern Recognition},
  pages     = {19447--19456},
  year      = {2024}
}

@inproceedings{yan2024multiscale,
  author    = {Zhiwen Yan and Weng Fei Low and Yu Chen and Gim Hee Lee},
  title     = {Multi-Scale 3D Gaussian Splatting for Anti-Aliased Rendering},
  booktitle = {Proceedings of the IEEE/CVF Conference on Computer Vision and Pattern Recognition},
  pages     = {20923--20931},
  year      = {2024}
}

@inproceedings{pumarola2021dnerf,
  author    = {Albert Pumarola and Enric Corona and Gerard Pons-Moll and Francesc Moreno-Noguer},
  title     = {D-{NeRF}: Neural Radiance Fields for Dynamic Scenes},
  booktitle = {Proceedings of the IEEE/CVF Conference on Computer Vision and Pattern Recognition},
  pages     = {10318--10327},
  year      = {2021}
}

@inproceedings{park2021nerfies,
  author    = {Keunhong Park and Utkarsh Sinha and Jonathan T. Barron and Sofien Bouaziz and Dan B. Goldman and Steven M. Seitz and Ricardo Martin-Brualla},
  title     = {Nerfies: Deformable Neural Radiance Fields},
  booktitle = {Proceedings of the IEEE/CVF International Conference on Computer Vision},
  pages     = {5865--5874},
  year      = {2021}
}

@inproceedings{wu2024fourDGS,
  author    = {Guanjun Wu and Taoran Yi and Jiemin Fang and Lingxi Xie and Xiaopeng Zhang and Wei Wei and Wenyu Liu and Qi Tian and Xinggang Wang},
  title     = {4D Gaussian Splatting for Real-Time Dynamic Scene Rendering},
  booktitle = {Proceedings of the IEEE/CVF Conference on Computer Vision and Pattern Recognition},
  pages     = {20310--20320},
  year      = {2024}
}

@article{fang2026DN2N,
  title={Editing 3d scenes via text prompts without retraining},
  author={Fang, Shuangkang and Wang, Yufeng and Tsai, Yi-Hsuan and Ding, Wenrui and Yang, Yi and Zhou, Shuchang and Yang, Ming-Hsuan},
  journal={IEEE Transactions on Visualization and Computer Graphics},
  year={2026},
  publisher={IEEE}
}

@inproceedings{yang2024deformable,
  author    = {Ziyi Yang and Xinyu Gao and Wen Zhou and Shaohui Jiao and Yuqing Zhang and Xiaogang Jin},
  title     = {Deformable 3D Gaussians for High-Fidelity Monocular Dynamic Scene Reconstruction},
  booktitle = {Proceedings of the IEEE/CVF Conference on Computer Vision and Pattern Recognition},
  pages     = {20331--20341},
  year      = {2024}
}

@inproceedings{kerr2023lerf,
  author    = {Justin Kerr and Chung Min Kim and Ken Goldberg and Angjoo Kanazawa and Matthew Tancik},
  title     = {{LERF}: Language Embedded Radiance Fields},
  booktitle = {Proceedings of the IEEE/CVF International Conference on Computer Vision},
  pages     = {19729--19739},
  year      = {2023}
}

@inproceedings{qin2024langsplat,
  author    = {Minghan Qin and Wanhua Li and Jiawei Zhou and Haoqian Wang and Hanspeter Pfister},
  title     = {LangSplat: 3D Language Gaussian Splatting},
  booktitle = {Proceedings of the IEEE/CVF Conference on Computer Vision and Pattern Recognition},
  pages     = {20051--20060},
  year      = {2024}
}

@inproceedings{zhou2024feature3dgs,
  author    = {Shijie Zhou and Haoran Chang and Sicheng Jiang and Zhiwen Fan and Zehao Zhu and Dejia Xu and Pradyumna Chari and Suya You and Zhangyang Wang and Achuta Kadambi},
  title     = {Feature 3DGS: Supercharging 3D Gaussian Splatting to Enable Distilled Feature Fields},
  booktitle = {Proceedings of the IEEE/CVF Conference on Computer Vision and Pattern Recognition},
  pages     = {21676--21685},
  year      = {2024}
}

@inproceedings{lu2024scaffold,
  author    = {Tao Lu and Mulin Yu and Linning Xu and Yuanbo Xiangli and Limin Wang and Dahua Lin and Bo Dai},
  title     = {Scaffold-{GS}: Structured 3D Gaussians for View-Adaptive Rendering},
  booktitle = {Proceedings of the IEEE/CVF Conference on Computer Vision and Pattern Recognition},
  pages     = {20654--20664},
  year      = {2024}
}

@inproceedings{li2024spacetime,
  author    = {Zhan Li and Zhang Chen and Zhong Li and Yi Xu},
  title     = {Spacetime Gaussian Feature Splatting for Real-Time Dynamic View Synthesis},
  booktitle = {Proceedings of the IEEE/CVF Conference on Computer Vision and Pattern Recognition},
  pages     = {8508--8520},
  year      = {2024}
}

@inproceedings{lee2024compact,
  author    = {Joo Chan Lee and Daniel Rho and Xiangyu Sun and Jong Hwan Ko and Eunbyung Park},
  title     = {Compact 3D Gaussian Representation for Radiance Field},
  booktitle = {Proceedings of the IEEE/CVF Conference on Computer Vision and Pattern Recognition},
  pages     = {21719--21728},
  year      = {2024}
}

@inproceedings{niedermayr2024compressed,
  author    = {Simon Niedermayr and Josef Stumpfegger and R{\"u}diger Westermann},
  title     = {Compressed 3D Gaussian Splatting for Accelerated Novel View Synthesis},
  booktitle = {Proceedings of the IEEE/CVF Conference on Computer Vision and Pattern Recognition},
  pages     = {10349--10358},
  year      = {2024}
}

@inproceedings{li2024dngaussian,
  author    = {Jiahe Li and Jiawei Zhang and Xiao Bai and Jin Zheng and Xin Ning and Jun Zhou and Lin Gu},
  title     = {{DNGaussian}: Optimizing Sparse-View 3D Gaussian Radiance Fields with Global-Local Depth Normalization},
  booktitle = {Proceedings of the IEEE/CVF Conference on Computer Vision and Pattern Recognition},
  pages     = {20775--20785},
  year      = {2024}
}

@inproceedings{yu2021pixelnerf,
  author    = {Alex Yu and Vickie Ye and Matthew Tancik and Angjoo Kanazawa},
  title     = {pixel{NeRF}: Neural Radiance Fields From One or Few Images},
  booktitle = {Proceedings of the IEEE/CVF Conference on Computer Vision and Pattern Recognition},
  pages     = {4578--4587},
  year      = {2021}
}

@inproceedings{chen2021mvsnerf,
  author    = {Anpei Chen and Zexiang Xu and Fuqiang Zhao and Xiaoshuai Zhang and Fanbo Xiang and Jingyi Yu and Hao Su},
  title     = {{MVSNeRF}: Fast Generalizable Radiance Field Reconstruction From Multi-View Stereo},
  booktitle = {Proceedings of the IEEE/CVF International Conference on Computer Vision},
  pages     = {14124--14133},
  year      = {2021}
}

@inproceedings{guedon2024sugar,
  author    = {Antoine Gu{\'e}don and Vincent Lepetit},
  title     = {{SuGaR}: Surface-Aligned Gaussian Splatting for Efficient 3D Mesh Reconstruction and High-Quality Mesh Rendering},
  booktitle = {Proceedings of the IEEE/CVF Conference on Computer Vision and Pattern Recognition},
  pages     = {5354--5363},
  year      = {2024}
}

@article{wang2004ssim,
  author  = {Zhou Wang and Alan C. Bovik and Hamid R. Sheikh and Eero P. Simoncelli},
  title   = {Image Quality Assessment: From Error Visibility to Structural Similarity},
  journal = {IEEE Transactions on Image Processing},
  volume  = {13},
  number  = {4},
  pages   = {600--612},
  year    = {2004},
  doi     = {10.1109/TIP.2003.819861}
}

@inproceedings{zhang2018lpips,
  author    = {Richard Zhang and Phillip Isola and Alexei A. Efros and Eli Shechtman and Oliver Wang},
  title     = {The Unreasonable Effectiveness of Deep Features as a Perceptual Metric},
  booktitle = {Proceedings of the IEEE Conference on Computer Vision and Pattern Recognition},
  pages     = {586--595},
  year      = {2018}
}

@article{fang2026pvdal,
  author  = {Shuangkang Fang and Yufeng Wang and Yi Yang and Weixin Xu and Heng Wang and Wenrui Ding and Shuchang Zhou},
  title   = {Progressive Volume Distillation with Active Learning for Efficient {NeRF} Architecture Conversion},
  journal = {International Journal of Computer Vision},
  volume  = {134},
  number  = {5},
  pages   = {228},
  year    = {2026},
  doi     = {10.1007/s11263-026-02815-1}
}

@inproceedings{fang2023pvd,
  author    = {Shuangkang Fang and Weixin Xu and Heng Wang and Yi Yang and Yufeng Wang and Shuchang Zhou},
  title     = {One Is All: Bridging the Gap between Neural Radiance Fields Architectures with Progressive Volume Distillation},
  booktitle = {Proceedings of the AAAI Conference on Artificial Intelligence},
  volume    = {37},
  number    = {1},
  pages     = {597--605},
  year      = {2023},
  doi       = {10.1609/aaai.v37i1.25135}
}

@inproceedings{fang2025nerfgs,
  author    = {Shuangkang Fang and I-Chao Shen and Takeo Igarashi and Yufeng Wang and Zesheng Wang and Yi Yang and Wenrui Ding and Shuchang Zhou},
  title     = {{NeRF} Is a Valuable Assistant for 3D Gaussian Splatting},
  booktitle = {Proceedings of the IEEE/CVF International Conference on Computer Vision},
  pages     = {26230--26240},
  year      = {2025}
}

@article{foroutan2024alternatives,
  author  = {Yalda Foroutan and Daniel Rebain and Kwang Moo Yi and Andrea Tagliasacchi},
  title   = {Evaluating Alternatives to {SfM} Point Cloud Initialization for Gaussian Splatting},
  journal = {arXiv preprint arXiv:2404.12547},
  year    = {2024}
}

@inproceedings{niemeyer2025radsplat,
  author    = {Michael Niemeyer and Fabian Manhardt and Marie-Julie Rakotosaona and Michael Oechsle and Daniel Duckworth and Rama Gosula and Keisuke Tateno and John Bates and Dominik Kaeser and Federico Tombari},
  title     = {{RadSplat}: Radiance Field-Informed Gaussian Splatting for Robust Real-Time Rendering with 900+ {FPS}},
  booktitle = {International Conference on 3D Vision},
  year      = {2025}
}

@article{malarz2025vdgs,
  author  = {Dawid Malarz and Weronika Smolak-Dy{\.z}ewska and Jacek Tabor and S{\l}awomir Tadeja and Przemys{\l}aw Spurek},
  title   = {Gaussian Splatting with {NeRF}-Based Color and Opacity},
  journal = {Computer Vision and Image Understanding},
  volume  = {251},
  pages   = {104273},
  year    = {2025},
  doi     = {10.1016/j.cviu.2024.104273}
}

@inproceedings{mihajlovic2024splatfields,
  author    = {Marko Mihajlovic and Sergey Prokudin and Siyu Tang and Robert Maier and Federica Bogo and Tony Tung and Edmond Boyer},
  title     = {{SplatFields}: Neural Gaussian Splats for Sparse 3D and 4D Reconstruction},
  booktitle = {European Conference on Computer Vision},
  pages     = {313--332},
  year      = {2024},
  doi       = {10.1007/978-3-031-72627-9_18}
}

@inproceedings{barthel2024decoder,
  author    = {Florian Barthel and Arian Beckmann and Wieland Morgenstern and Anna Hilsmann and Peter Eisert},
  title     = {Gaussian Splatting Decoder for 3D-Aware Generative Adversarial Networks},
  booktitle = {Proceedings of the IEEE/CVF Conference on Computer Vision and Pattern Recognition Workshops},
  pages     = {7963--7972},
  year      = {2024},
  doi       = {10.1109/CVPRW63382.2024.00794}
}

@article{yang2023unified,
  author  = {Xiaofeng Yang and Yiwen Chen and Cheng Chen and Chi Zhang and Yi Xu and Xulei Yang and Fayao Liu and Guosheng Lin},
  title   = {Learn to Optimize Denoising Scores for 3D Generation: A Unified and Improved Diffusion Prior on {NeRF} and 3D Gaussian Splatting},
  journal = {arXiv preprint arXiv:2312.04820},
  year    = {2023}
}

@article{morikawa2026leveraging,
  author  = {Mizuki Morikawa and Yuta Shimizu and Chunyu Li and Yusuke Monno and Masatoshi Okutomi},
  title   = {Leveraging {NeRF}-Rendered Images for 3D Gaussian Splatting},
  journal = {arXiv preprint arXiv:2606.09034},
  year    = {2026}
}

@inproceedings{xu2022pointnerf,
  author    = {Qiangeng Xu and Zexiang Xu and Julien Philip and Sai Bi and Zhixin Shu and Kalyan Sunkavalli and Ulrich Neumann},
  title     = {Point-{NeRF}: Point-Based Neural Radiance Fields},
  booktitle = {Proceedings of the IEEE/CVF Conference on Computer Vision and Pattern Recognition},
  pages     = {5438--5448},
  year      = {2022}
}

@inproceedings{chen2023mobilenerf,
  author    = {Zhiqin Chen and Thomas Funkhouser and Peter Hedman and Andrea Tagliasacchi},
  title     = {Mobile{NeRF}: Exploiting the Polygon Rasterization Pipeline for Efficient Neural Field Rendering on Mobile Architectures},
  booktitle = {Proceedings of the IEEE/CVF Conference on Computer Vision and Pattern Recognition},
  pages     = {16569--16578},
  year      = {2023}
}

@inproceedings{hedman2021snereg,
  author    = {Peter Hedman and Pratul P. Srinivasan and Ben Mildenhall and Jonathan T. Barron and Paul Debevec},
  title     = {Baking Neural Radiance Fields for Real-Time View Synthesis},
  booktitle = {Proceedings of the IEEE/CVF International Conference on Computer Vision},
  pages     = {5875--5884},
  year      = {2021}
}

@inproceedings{yu2021plenoctrees,
  author    = {Alex Yu and Ruilong Li and Matthew Tancik and Hao Li and Ren Ng and Angjoo Kanazawa},
  title     = {{PlenOctrees} for Real-Time Rendering of Neural Radiance Fields},
  booktitle = {Proceedings of the IEEE/CVF International Conference on Computer Vision},
  pages     = {5752--5761},
  year      = {2021}
}

@inproceedings{reiser2021kilonerf,
  author    = {Christian Reiser and Songyou Peng and Yiyi Liao and Andreas Geiger},
  title     = {Kilo{NeRF}: Speeding Up Neural Radiance Fields with Thousands of Tiny {MLPs}},
  booktitle = {Proceedings of the IEEE/CVF International Conference on Computer Vision},
  pages     = {14335--14345},
  year      = {2021}
}

@inproceedings{turki2024hybrid,
  author    = {Haithem Turki and Vasu Agrawal and Samuel Rota Bul\`o and Lorenzo Porzi and Peter Kontschieder and Deva Ramanan and Michael Zollh{\"o}fer and Christian Richardt},
  title     = {Hybrid{NeRF}: Efficient Neural Rendering via Adaptive Volumetric Surfaces},
  booktitle = {Proceedings of the IEEE/CVF Conference on Computer Vision and Pattern Recognition},
  pages     = {19647--19656},
  year      = {2024}
}

@inproceedings{cheng2024gaussianpro,
  author    = {Kai Cheng and Xiaoxiao Long and Kaizhi Yang and Yao Yao and Wei Yin and Yuexin Ma and Wenping Wang and Xuejin Chen},
  title     = {{GaussianPro}: 3D Gaussian Splatting with Progressive Propagation},
  booktitle = {Proceedings of the 41st International Conference on Machine Learning},
  series    = {Proceedings of Machine Learning Research},
  volume    = {235},
  pages     = {8123--8140},
  year      = {2024}
}

@inproceedings{niemeyer2022regnerf,
  author    = {Michael Niemeyer and Jonathan T. Barron and Ben Mildenhall and Mehdi S. M. Sajjadi and Andreas Geiger and Noha Radwan},
  title     = {Reg{NeRF}: Regularizing Neural Radiance Fields for View Synthesis from Sparse Inputs},
  booktitle = {Proceedings of the IEEE/CVF Conference on Computer Vision and Pattern Recognition},
  pages     = {5480--5490},
  year      = {2022}
}

@inproceedings{deng2022dsnerf,
  author    = {Kangle Deng and Andrew Liu and Jun-Yan Zhu and Deva Ramanan},
  title     = {Depth-Supervised {NeRF}: Fewer Views and Faster Training for Free},
  booktitle = {Proceedings of the IEEE/CVF Conference on Computer Vision and Pattern Recognition},
  pages     = {12882--12891},
  year      = {2022}
}

@inproceedings{kim2022infonerf,
  author    = {Mijeong Kim and Seonguk Seo and Bohyung Han},
  title     = {Info{NeRF}: Ray Entropy Minimization for Few-Shot Neural Volume Rendering},
  booktitle = {Proceedings of the IEEE/CVF Conference on Computer Vision and Pattern Recognition},
  pages     = {12912--12921},
  year      = {2022}
}

@inproceedings{yang2023freenerf,
  author    = {Jiawei Yang and Marco Pavone and Yue Wang},
  title     = {Free{NeRF}: Improving Few-Shot Neural Rendering with Free Frequency Regularization},
  booktitle = {Proceedings of the IEEE/CVF Conference on Computer Vision and Pattern Recognition},
  pages     = {8254--8263},
  year      = {2023}
}

@inproceedings{jain2021dietnerf,
  author    = {Ajay Jain and Matthew Tancik and Pieter Abbeel},
  title     = {Putting {NeRF} on a Diet: Semantically Consistent Few-Shot View Synthesis},
  booktitle = {Proceedings of the IEEE/CVF International Conference on Computer Vision},
  pages     = {5885--5894},
  year      = {2021}
}

@article{jung2024raings,
  author  = {Jaewoo Jung and Jisang Han and Honggyu An and Jiwon Kang and Seonghoon Park and Seungryong Kim},
  title   = {Relaxing Accurate Initialization Constraint for 3D Gaussian Splatting},
  journal = {arXiv preprint arXiv:2403.09413},
  year    = {2024}
}

@inproceedings{huang2024twodgs,
  author    = {Binbin Huang and Zehao Yu and Anpei Chen and Andreas Geiger and Shenghua Gao},
  title     = {2D Gaussian Splatting for Geometrically Accurate Radiance Fields},
  booktitle = {ACM SIGGRAPH 2024 Conference Papers},
  year      = {2024},
  doi       = {10.1145/3641519.3657428}
}

@article{yu2024gof,
  author  = {Zehao Yu and Torsten Sattler and Andreas Geiger},
  title   = {Gaussian Opacity Fields: Efficient Adaptive Surface Reconstruction in Unbounded Scenes},
  journal = {ACM Transactions on Graphics},
  volume  = {43},
  number  = {6},
  pages   = {271:1--271:13},
  year    = {2024},
  doi     = {10.1145/3687937}
}

@inproceedings{kheradmand2024mcmc,
  author    = {Shakiba Kheradmand and Daniel Rebain and Gopal Sharma and Weiwei Sun and Yang-Che Tseng and Hossam Isack and Abhishek Kar and Andrea Tagliasacchi and Kwang Moo Yi},
  title     = {3D Gaussian Splatting as Markov Chain Monte Carlo},
  booktitle = {Advances in Neural Information Processing Systems},
  volume    = {37},
  pages     = {80965--80986},
  year      = {2024}
}

@article{radl2024stopthepop,
  author  = {Lukas Radl and Michael Steiner and Mathias Parger and Alexander Weinrauch and Bernhard Kerbl and Markus Steinberger},
  title   = {{StopThePop}: Sorted Gaussian Splatting for View-Consistent Real-Time Rendering},
  journal = {ACM Transactions on Graphics},
  volume  = {43},
  number  = {4},
  pages   = {64:1--64:17},
  year    = {2024},
  doi     = {10.1145/3658187}
}

@inproceedings{zhang2024pixelgs,
  author    = {Zheng Zhang and Wenbo Hu and Yixing Lao and Tong He and Hengshuang Zhao},
  title     = {Pixel-{GS}: Density Control with Pixel-Aware Gradient for 3D Gaussian Splatting},
  booktitle = {European Conference on Computer Vision},
  pages     = {326--342},
  year      = {2024},
  doi       = {10.1007/978-3-031-72655-2_19}
}

@inproceedings{xie2025hashgs,
  author    = {Yijia Xie and Yuhang Lin and Laijian Li and Lina Liu and Xiaobin Wei and Yong Liu and Jiajun Lv},
  title     = {Hash-{GS}: Anchor-Based 3D Gaussian Splatting with Multi-Resolution Hash Encoding for Efficient Scene Reconstruction},
  booktitle = {IEEE International Conference on Robotics and Automation},
  pages     = {13964--13971},
  year      = {2025},
  doi       = {10.1109/ICRA55743.2025.11128324}
}

@inproceedings{charatan2024pixelsplat,
  author    = {David Charatan and Sizhe Lester Li and Andrea Tagliasacchi and Vincent Sitzmann},
  title     = {pixelSplat: 3D Gaussian Splats from Image Pairs for Scalable Generalizable 3D Reconstruction},
  booktitle = {Proceedings of the IEEE/CVF Conference on Computer Vision and Pattern Recognition},
  pages     = {19457--19467},
  year      = {2024}
}

@inproceedings{szymanowicz2024splatter,
  author    = {Stanislaw Szymanowicz and Chrisitian Rupprecht and Andrea Vedaldi},
  title     = {Splatter Image: Ultra-Fast Single-View 3D Reconstruction},
  booktitle = {Proceedings of the IEEE/CVF Conference on Computer Vision and Pattern Recognition},
  pages     = {10208--10217},
  year      = {2024}
}

@inproceedings{zou2024triplane,
  author    = {Zi-Xin Zou and Zhipeng Yu and Yuan-Chen Guo and Yangguang Li and Ding Liang and Yan-Pei Cao and Song-Hai Zhang},
  title     = {Triplane Meets Gaussian Splatting: Fast and Generalizable Single-View 3D Reconstruction with Transformers},
  booktitle = {Proceedings of the IEEE/CVF Conference on Computer Vision and Pattern Recognition},
  pages     = {10324--10335},
  year      = {2024}
}

@inproceedings{chen2024mvsplat,
  author    = {Yuedong Chen and Haofei Xu and Chuanxia Zheng and Bohan Zhuang and Marc Pollefeys and Andreas Geiger and Tat-Jen Cham and Jianfei Cai},
  title     = {{MVSplat}: Efficient 3D Gaussian Splatting from Sparse Multi-View Images},
  booktitle = {European Conference on Computer Vision},
  pages     = {370--386},
  year      = {2024}
}

@inproceedings{tang2024lgm,
  author    = {Jiaxiang Tang and Zhaoxi Chen and Xiaokang Chen and Tengfei Wang and Gang Zeng and Ziwei Liu},
  title     = {{LGM}: Large Multi-View Gaussian Model for High-Resolution 3D Content Creation},
  booktitle = {European Conference on Computer Vision},
  pages     = {1--18},
  year      = {2024},
  doi       = {10.1007/978-3-031-73235-5_1}
}

@inproceedings{xu2025depthsplat,
  author    = {Haofei Xu and Songyou Peng and Fangjinhua Wang and Hermann Blum and Daniel Barath and Andreas Geiger and Marc Pollefeys},
  title     = {{DepthSplat}: Connecting Gaussian Splatting and Depth},
  booktitle = {Proceedings of the IEEE/CVF Conference on Computer Vision and Pattern Recognition},
  pages     = {16453--16463},
  year      = {2025}
}

@inproceedings{zhang2025sogs,
  author    = {Jiahui Zhang and Fangneng Zhan and Ling Shao and Shijian Lu},
  title     = {{SOGS}: Second-Order Anchor for Advanced 3D Gaussian Splatting},
  booktitle = {Proceedings of the IEEE/CVF Conference on Computer Vision and Pattern Recognition},
  pages     = {11167--11176},
  year      = {2025}
}

@article{fang2026dropansh,
  author  = {Shuangkang Fang and I-Chao Shen and Xuanyang Zhang and Zesheng Wang and Yufeng Wang and Wenrui Ding and Gang Yu and Takeo Igarashi},
  title   = {Dropping Anchor and Spherical Harmonics for Sparse-view Gaussian Splatting},
  journal = {arXiv preprint arXiv:2602.20933},
  year    = {2026}
}
